\documentclass[11pt,a4paper]{article}
\usepackage[hyperindex,breaklinks]{hyperref}
\usepackage{breakurl}
\usepackage[]{acl2019}
\usepackage{times}
\usepackage{latexsym}

\aclfinalcopy 

\title{Shared Task: Lexical Semantic Change Detection in German}

\author{Adnan Ahmad, Kiflom Desta, Fabian Lang, Dominik Schlechtweg \\
  Institute for Natural Language Processing, University of Stuttgart\\
  {\tt adnan.sust@outlook.com}, {\tt desta7883@gmail.com},\\ {\tt f.lang.1993@web.de}, {\tt schlecdk@ims.uni-stuttgart.de}  }

\date{}

\begin{document}
\maketitle
\begin{abstract}
Recent NLP architectures have illustrated in various ways how semantic change can be captured across time and domains. However, in terms of evaluation there is a lack of benchmarks to compare the performance of these systems against each other. We present the results of the first shared task on unsupervised lexical semantic change detection (LSCD) in German based on the evaluation framework proposed by \citet{DBLP:journals/corr/abs-1906-02979}.
\end{abstract}

\section{Introduction}
Natural languages evolve and words have always been subject to semantic change over time \cite{traugott2001regularity}. With the rise of large digitized text resources recent NLP technologies have made it possible to capture such change with vector space models \citep*{pennington2014glove,rudolph2017dynamic,bengio2003neural,rosenfeld-erk-2018-deep}, topic models \citep*{wang2006topics,lau2012word,frermann2016bayesian}, and sense clustering models \citep{mitra2015automatic}. However, many approaches for detecting LSC differ profoundly from each other and therefore drawing comparisons between them can be challenging \citep{2018arXiv181106278T}. Not only do architectures for detecting LSC vary, their performance is also often evaluated without access to evaluation data or too sparse data sets. In cases where evaluation data is available, oftentimes LSCD systems are not evaluated on the same data set which hinders the research community to draw comparisons. 

For this reason we report the results of the first shared task on unsupervised lexical semantic change detection in German\footnote{\url{https://codalab.lri.fr/competitions/560}} that is based on an annotated data set to guarantee objective reasoning throughout different approaches. The task was organized as part of the seminar 'Lexical Semantic Change Detection' at the IMS Stuttgart in the summer term of 2019.\footnote{\url{https://www.f05.uni-stuttgart.de/informatik/dokumente/Seminare/Seminare-SS_2019/04_LexicalSemantics.pdf}}

\section{Task}
The goal of the shared task was to create an architecture to detect semantic change and to rank words according to their degree of change between two different time periods. Given two corpora C\textsubscript{a} and C\textsubscript{b}, the target words had to be ranked according to their degree of lexical semantic change between C\textsubscript{a} and C\textsubscript{b} as annotated by human judges. A competition was set up on Codalab and teams mostly consisting of 2 people were formed to take part in the task. There was one group consisting of 3 team members and two individuals who entered the task on their own. In total there were 12 LSCD systems participating in the shared task.

The shared task was divided into three phases, i.e., development, testing and analysis phase. In the development phase each team implemented a first version of their model based on a trial data set and submitted it subsequently. In the testing phase the testing data was made public and participants applied their models to the test data with a restriction of possible result uploads to 30. The leaderboard was public at all times. Eventually, the analysis phase was entered and the models of the testing phase were evaluated in terms of the predictions they made and parameters could be tuned further. The models and results will be discussed in detail in sections 7 and 8.

\section{Corpora}
The task, as framed above, requires to detect the semantic change between two corpora. The two corpora used in the shared task correspond to the diachronic corpus pair from \citet{DBLP:journals/corr/abs-1906-02979}: DTA18 and DTA19.\footnote{\url{https://www.ims.uni-stuttgart.de/forschung/ressourcen/korpora/wocc/}} They consist of subparts of DTA corpus \citep{dta2017} which is a freely available lemmatized, POS-tagged and spelling-normalized diachronic corpus of German containing texts from the 16th to the 20th century. DTA18 contains 26 million sentences published between 1750-1799 and DTA19 40 million between 1850-1899. The corpus version used in the task has the following format: "year [tab] lemma1 lemma2 lemma3 ...".

\section{Evaluation}
The Diachronic Usage Relatedness (DURel) gold standard data set includes 22 target words and their varying degrees of semantic change \cite{schlechtweg2018diachronic}. For each of these target words a random sample of use pairs from the DTA corpus was retrieved and annotated. The annotators were required to rate the pairs according to their semantic relatedness on a scale from 1 to 4 (unrelated - identical meanings) for two time periods. The average Spearman's $\rho$ between the five annotators was 0.66 for 1,320 use paris. The resulting word ranking of the DURel data set is determined by the mean usage relatedness across two time periods and is used as the benchmark to compare the models’ performances in the shared task. 

\subsection{Metric}
The output of a system with the target words in the predicted order is compared to the gold ranking of the DURel data set. As the metric to assess how well the model's output fits the gold ranking Spearman's $\rho$ was used. The higher Spearman's rank-order correlation the better the system's performance.

\subsection{Baselines}
Models were compared to two baselines for the shared task:
\begin{enumerate}
    \item log-transformed normalized frequency difference (FD)
    \item count vectors with column intersection and cosine distance (CNT + CI + CD)
\end{enumerate}
The window size for CNT + CI + CD was 10. Find more information on these models in \citet{DBLP:journals/corr/abs-1906-02979}.

\begin{table*}[]
\begin{tabular}{|l|lll|l|l|}
\hline
\textbf{Team} & \multicolumn{1}{l|}{\textbf{Space}} & \multicolumn{1}{l|}{\textbf{Align}} & \textbf{Measure} & \textbf{Spearman} & \textbf{Comment} \\ \hline
sorensbn & \multicolumn{1}{l|}{SGNS} & \multicolumn{1}{l|}{OP} & CD & .854 & Noise-aware alignment \\ \hline
tidoe & \multicolumn{1}{l|}{SGNS} & \multicolumn{1}{l|}{OP} & CD & .811 & Binarized matrices \\ \hline
in vain & \multicolumn{1}{l|}{SGNS} & \multicolumn{1}{l|}{VI} & CD & .802 &  \\ \hline
Evilly & \multicolumn{1}{l|}{SGNS} & \multicolumn{1}{l|}{OP} & CD & .730 & Frequency-driven OP alignment \\ \hline
DAF & \multicolumn{1}{l|}{fastText} & \multicolumn{1}{l|}{OP} & CD & .570 &  \\ \hline
SnakesOnAPlane & \multicolumn{1}{l|}{SGNS} & \multicolumn{1}{l|}{OP} & CD / JSD & .565 / .561 & Measure comparison \\ \hline
TeamKulkarni15 & \multicolumn{1}{l|}{SGNS} & \multicolumn{1}{l|}{OP} & CD & .540 & Local alignment with KNN \\ \hline
Bashmaistori & \multicolumn{1}{l|}{PPMI} & \multicolumn{1}{l|}{WI} & CD & .511 &  \\ \hline
Baseline 2 & \multicolumn{1}{l|}{CNT} & \multicolumn{1}{l|}{CI} & CD & .486 & \\ \hline
giki & \multicolumn{1}{l|}{PPMI} & \multicolumn{1}{l|}{CI} & CD & .432 &  \\ \hline
Edu-Phil & \multicolumn{1}{l|}{fastText} & \multicolumn{1}{l|}{OP} & CD & .381 &  \\ \hline
orangefoxes & \multicolumn{1}{l|}{SGNS} & \multicolumn{1}{l|}{-} & CD & .121 & DiffTime \\ \hline
Loud Whisper & \multicolumn{1}{l|}{-} & \multicolumn{1}{l|}{-} & - & .092 & Graph-based approach \\ \hline
Baseline 1 & \multicolumn{1}{l|}{-} & \multicolumn{1}{l|}{-} & FD & .019 &  \\ \hline
\end{tabular}
\caption{Shared task - Overview of participating systems. Legend: Space = Semantic space; Align = Alignment method; Measure = Distance Measure for LSC detection.
\textbf{Note:} The best result either obtained in testing or analysis phase is reported.}
\label{table:tab1}
\end{table*}

\section{Participating Systems}
Participants mostly rely on the models compared in \citet{DBLP:journals/corr/abs-1906-02979} and apply modifications to improve them.\footnote{Find implementations at \url{https://github.com/Garrafao/LSCDetection}.} In particular, most teams make use of skip-gram with negative sampling (SGNS) based on \citet{mikolov2013efficient} to learn the semantic spaces of the two time periods and orthogonal procrustes (OP) to align these vector spaces, similar to the approach by \citet{hamilton-etal-2016-diachronic}. Different meaning representations such as sense clusters are used as well. As measure to detect the degree of LSC all teams except one choose cosine distance (CD). This team uses Jensen-Shannon distance (JSD) instead, which computes the distance between probability distributions \cite{lin1991divergence}.

The models of each team will be briefly introduced in this section. \paragraph{sorensbn} Team sorensbn makes use of SGNS + OP + CD to detect LSC. They use similar hyperparameters as in \citet{DBLP:journals/corr/abs-1906-02979} to tune the SGNS model. They use an open-sourced noise-aware implementation to improve the OP alignment \cite{yehezkel-lubin-etal-2019-aligning}. \paragraph{tidoe} Team tidoe builds on SGNS + OP + CD, but they add a transformation step to receive binarized representations of matrices \cite{DBLP:journals/corr/FaruquiTYDS15}. This step is taken to counter the bias that can occur in vector-space models based on frequencies \citep{dubossarsky2017}. \paragraph{in vain} The team applies a model based on SGNS with vector initialization alignment and cosine distance (SGNS + VI + CD). Vector initialization is an alignment strategy where the vector space learning model for $t_2$ is initialized with the vectors from $t_1$ \cite{kim-etal-2014-temporal}. Since SGNS + VI + OP does not perform as well as other models in \citet{DBLP:journals/corr/abs-1906-02979}, they alter the vector initialization process by initializing on the complete model instead of only the word matrix of $t_1$ to receive improved results. \paragraph{Evilly} In line with previous approaches, team Evilly builds upon SGNS + OP + CD. They alter the OP step by using only high-frequency words for alignment. \paragraph{DAF} Team DAF uses an architecture based on learning vectors with fastText, alignment with unsupervised and supervised variations of OP, and CD, using the MUSE package\footnote{Find package at: \url{https://github.com/facebookresearch/MUSE}} \citep*{conneau2017word,joulin2016fasttext}. For the supervised alignment stop words are used. The underlying assumption is that stop words serve as functional units of language and their usage should be consistent over time. \paragraph{SnakesOnAPlane}
The team learns vector spaces with count vectors, positive pointwise mutual information (PPMI), SGNS and uses column intersection (CI) and OP as alignment techniques where applicable. Then they compare two distance measures (CD and JSD) for the different models CNT + CI, PPMI + CI and SGNS + OP to identify which measure performs better for these models. They also experiment with different ways to remove negative values from SGNS vectors, which is needed for JSD. \paragraph{TeamKulkarni15} TeamKulkarni15 uses SGNS + OP + CD with the modification of local alignment with k nearest neighbors, since other models often use global alignment that can be prone to noise \cite{DBLP:journals/corr/KulkarniAPS14a}. \paragraph{Bashmaistori} They use word injection (WI) alignment on PPMI vectors with CD. This approach avoids the complex alignment procedure for embeddings and is applicable to embeddings and count-based methods. They compare two implementations of word injection \citep*{DBLP:journals/corr/abs-1906-01688,DBLP:journals/corr/abs-1906-02979} as these showed different results on different data sets.  \paragraph{giki} Team giki uses PPMI + CI + CD to detect LSC. They state that a word sense is determined by its context, but relevant context words can also be found outside a predefined window. Therefore, they use tf-idf to select relevant context \cite{ramos2003using}. \paragraph{Edu-Phil} Similar to team DAF they also use fastText + OP + CD. Their hypothesis is that fastText may increase the performance for less frequent words in the corpus since generating word embeddings in fasttext is based on character n-grams. \paragraph{orangefoxes} They use the model by \citet{rosenfeld-erk-2018-deep} which is based on SGNS, but avoids alignment by treating time as a vector that may be combined with word vectors to get time-specific word vectors. \paragraph{Loud Whisper} Loud Whisper base their approach on \citet{mitra2015automatic} which is a graph-based sense clustering model. They process the data set to receive bigrams, create a co-occurence graph representation and after clustering assess the type of change per word by comparing the results against an intersection table. Their motivation is not only to use a graph-based approach, but to extend the approach by enabling change detection for all parts of speech as opposed to the original model.

\section{Results and Discussion}
Table \ref{table:tab1} shows the results of the shared task. All teams receive better results than baseline 1 (FD), of which a total of 8 teams outperform baseline 2 (CNT + CI + CD). The 4 top scores with $\rho$ $>$ 0.7 are either modified versions of SGNS + OP + CD or use SGNS + VI + CD. The following 4 scores in the range of 0.5 $<$ $\rho$ $<$ 0.6 are generated by the models fastText + OP + CD, SGNS + OP + CD/JSD, and PPMI + WI + CD. 

Contrary to the results by \citet{DBLP:journals/corr/abs-1906-02979} the modified version of vector initialization shows high performance similar to OP alignment, as previously reported by \citet{hamilton-etal-2016-diachronic}. Some modifications to the SGNS + OP + CD approach are able to yield better results than others, e.g. noise-aware alignment and binarized matrices as compared to frequency-driven OP alignment or local alignment with KNN. Team SnakesOnAPlane compare two distance measures and their results show that JSD ($\rho$ $=$ .561) performs minimally worse than CD ($\rho$ $=$ .565) as the semantic change measure for their model.

The overall best-performing model is Skip-Gram with orthogonal alignment and cosine distance (SGNS + OP + CD) with similar hyperparameters as in the model architecture described previously \citep{DBLP:journals/corr/abs-1906-02979}. Said architecture was used as the basis for the two best performing models. Team tidoe reports that binarizing matrices leads to a generally worse performance ($\rho$ $=$ .811) compared to the unmodified version of SGNS + OP + CD ($\rho$ $=$ 0.9). The noise aware alignment approach applied by team sorensbn obtains a higher score ($\rho$ $=$ .854) compared to the result reported by tidoe, but is unable to exceed the performance of the unmodified SNGS + OP + CD for the same set of hyperparameters (window size = 10, negative sampling = 1; subsampling = None). Of the 8 scores above the second baseline, 5 use an architecture that builds upon SGNS + OP + CD. Whereas in the lower score segment $\rho$ $<$ 0.5 none of the models use SGNS + OP + CD. These findings are in line with the results reported by \citet{DBLP:journals/corr/abs-1906-02979}, however the overall best results are lower in this shared task, which is expected from the smaller number of parameter combinations explored.  Additionally, in the shared task the objective was to report the best score and not to calculate the mean which makes it more difficult to compare the robustness of the models presented here.

\bibliography{acl2019}
\bibliographystyle{acl_natbib}

\end{document}